\documentclass{article}

\usepackage{PRIMEarxiv}

\usepackage[utf8]{inputenc} % allow utf-8 input
\usepackage[T1]{fontenc}    % use 8-bit T1 fonts
\usepackage{hyperref}       % hyperlinks
\usepackage{url}            % simple URL typesetting
\usepackage{booktabs}       % professional-quality tables
\usepackage{amsfonts}       % blackboard math symbols
\usepackage{nicefrac}       % compact symbols for 1/2, etc.
\usepackage{microtype}      % microtypography
\usepackage{lipsum}
\usepackage{fancyhdr}       % header
\usepackage{graphicx}       % graphics
\graphicspath{{media/}}     % organize your images and other figures under media/ folder
\usepackage{array}
\usepackage{booktabs} % For better-looking tables
\usepackage{tabularx}
\usepackage[german=quotes]{csquotes}
\usepackage[backend=biber,style=ieee]{biblatex}
\title{Phoenix: Open-Source Language Adaption for Direct Preference Optimization}
\author{
  Matthias Uhlig \\
  Hochschule Ansbach \\
  \texttt{uhlig19498@hs-ansbach.de} \\
   \And
 Sigurd Schacht \\
  Hochschule Ansbach \\
  \texttt{sigurd.schacht@hs-ansbach.de} \\
  \And
  Sudarshan Kamath Barkur \\
  Hochschule Ansbach \\
  \texttt{s.kamath-barkur@hs-ansbach.de} \\
}

\begin{document}
\maketitle

\begin{abstract}
Large language models have gained immense importance in recent years and have demonstrated outstanding results in solving various tasks. However, despite these achievements, many questions remain unanswered in the context of large language models. Besides the optimal use of the models for inference and the alignment of the results to the desired specifications, the transfer of models to other languages is still an underdeveloped area of research. The recent publication of models such as Llama-2 and Zephyr has provided new insights into architectural improvements and the use of human feedback. However, insights into adapting these techniques to other languages remain scarce.  In this paper, we build on latest improvements and apply the Direct Preference Optimization(DPO) approach to the German language. The model is available at \url{https://huggingface.co/DRXD1000/Phoenix}.
\end{abstract}

% keywords can be removed

\section{Introduction}
\label{Introduction}

Large Language Models like GPT-3  \cite{brown_language_nodate}, Llama-2 \cite{touvron_llama_2023} or Mistral \cite{Jiang2023-qd-mistral} demonstrate very good results in solving a wide range of tasks and form good foundation models for adaptation to chat models \cite{taori_alpaca_nodate, tunstall_zephyr_2023} or other domain-specific applications. These and other open-source models benefit from a wide range of technological improvements. Structural improvements such as Grouped-Query Attention \cite{ainslie_gqa_2023} or Flash-Attention \cite{dao_flashattention_2022,dao_flashattention-2_2023} for example were able to drastically reduce both the calculation time and the memory overhead of the attention block. Furthermore, high-quality closed-source models such as GPT-4\cite{openai_gpt-4_2023}, ChatGPT \cite{chatgpt-introduction}, Claude 2 \cite{claude_2_anthropic} and Crowdsourcing efforts  \cite{kopf_openassistant_2023} allowed the creation of high-quality datasets. These models act as teachers, distilling the information and allowing the creation of student models trained with this distilled information.

However, despite this highly diverse field of research, which, in addition to the points already mentioned, also deals with inference optimization i.a. \cite{dettmers_llmint8_2022,frantar_gptq_2023,frantar_sparsegpt_2023,lin_awq_2023} and improving the availability of those models to the general public and low-resource researchers \cite{hu_lora_2021,dettmers_qlora_2023}, there still remains one element, the language, which has continued to be highly homogeneous. All previously mentioned enhancements to data, models, or pipelines aim to improve the results on English-based benchmarks, even though there are only around 300 million native speakers around the globe. The focus on those standard benchmarks through which models are compared results in a drastic under-representation of languages other than English. Furthermore, it also influences the selection process regarding the pretraining corpus. Although there are around 100 million German native speakers, the Llama-2 model family corpus only contains 0.17\% of German texts \cite{touvron_llama_2023}. As a result, the models are under-trained for languages other than English.

\section{Related Work}
\label{Related Work}

Training the model in multiple languages is possible. This can be achieved using pre-training the model first with the corpus consisting of multiple languages, followed by fine-tuning for downstream tasks. For example, Google trained the mT5 model from scratch over 101 Languages\cite{Xue2020-nv}, Meta trained the XGLM model with specific tasks for more than 20 languages \cite{Lin2021-ok} and the BigScience Workshop trained BLOOM Model with over 46 languages \cite{BigScience_Workshop2022-ga}. However, the large number of languages in these models means that the performance in a specific language declines sharply and does not come close to the quality of monolingual models \cite{minixhofer_wechsel_2022}. To solve this problem, approaches have already been developed. For example, the WECHSEL approach \cite {minixhofer_wechsel_2022} replaces the tokenizer used with a new tokenizer(for the target language) such that the new tokens are semantically similar. The CLP-Transfer approach \cite{ostendorff_efficient_2023} uses a smaller, resource-efficient model in the target language. This smaller model, along with the source model, helps initialize the token embeddings of the larger model by leveraging the shared vocabulary of both languages. The rest of the weights are retained from the source language model. However, these methods require an extremely high computing effort that most organizations cannot cope with. Concerning the variants, only a GPT-2 model of \cite{minixhofer_wechsel_2022} and a Bloom-6.7B model of \cite{ostendorff_efficient_2023} exist. Due to the now obsolete architectures of these models, neither Flash Attention nor Grouped Query Attention can be used with them. 

Since models learn almost their entire knowledge in pretraining \cite{zhou_lima_2023}, the LAION team \cite{Plüster_2023} pursued another approach by further training current models using a large language-specific text corpus in the style of casual language modeling. With an additional 65 billion training tokens, these models can outperform their original English counterparts in German tasks and partially outperform the original models in English tasks \cite{Plüster_2023}. Besides this, the LAION team was also able to translate known instruction data sets into German with the help of the OpenAI API to train German chat models.

While Supervised Finetuning(SFT) provides a good approach to teach a model the conversation or response pattern of humans or other, mostly larger, models, they tend to give false or toxic answers \cite{ganguli_red_2022}. To improve the alignment of models and increase the safety and helpfulness of LLMs, new training methods using reinforcement learning have been applied. This approach also known as Reinforcement Learning with Human Feedback(RLHF) has been used among others by the OpenAI Team to train InstructGPT \cite{ouyang_training_2022} and Meta to improve Llama-2 \cite{touvron_llama_2023}. InstructGPT thereby outperformed GPT-3 175B significantly in Human Evaluation although the model only has 1.3B parameters. The use of the Proximal Policy Optimization(PPO) \cite{Schulman2017-jb} has been the State of the Art Pipeline to to train a RLHF Model \cite{Ziegler2019-pa}, but the need for a separate reward model added additional complexity to the process. A newer approach, Direct Preference Optimization(DPO) \cite{rafailov_direct_2023}, made this step dispensable and allowed a direct optimization of the SFT Model by comparing the scores for a chosen and rejected answer and adapting the weights of the model to maximize the rewards.  

Since the release of the LAION models \cite{Plüster_2023} further improvements have become available that could not be included in their training.  Neftune \cite{Jain2023-ho}, for example, has significantly improved the quality of instruction tuning. Furthermore, the work of HuggingFace for the creation of the Zephyr model \cite{tunstall_zephyr_2023} using Direct Preference Optimization(DPO) \cite{rafailov_direct_2023} enabled significant progress in the alignment of language models and thus made it possible to compete with Zephyr 7B against models one order of magnitude bigger. To improve the work of the LAION team \cite{Plüster_2023} and to further promote the use of LLMs for the German language, we extend the approach pursued so far and optimize the Mistral \cite{Jiang2023-qd-mistral} adaptation of the LAION team\cite{Plüster_2023} by extended Supervised Finetuning and the use of DPO. 

\section{Method}

While the HuggingFace Hub already offers a variety of translated German instruction datasets, like variants of Share-GPT, Alpaca, and Evol-Instruct provided by FreedomAI\cite{Chen_MultilingualSIFT_Multilingual_Supervised_2023}, Open-Platypus from the LAION team\cite{Plüster_2023} and Dolly 15K \cite{DatabricksBlog2023DollyV2} translated to German by ourselves \cite{german_dolly-15k}, most of the entries consist of single turn conversations which limit their suitability for multi-turn chatbots.

\subsection{Translation with open-source tools}

The HuggingFace team behind the Zephyr model \cite{tunstall_zephyr_2023} has chosen a different dataset and used a cleaned and filtered versions of data from the Tsinghua University \cite{ding2023ultrachat} for instruction data and UltraFeedback dataset\cite{cui2023ultrafeedback} for DPO. While the adaptions of these datasets offer a high diversity of multi-turn conversations and feedback data, translating them with traditional API services would be a huge financial investment. Together, the Ultrachat version from HuggingFace\cite{tunstall_zephyr_2023} and the post-processed version of the Ultrafeedback dataset from Argilla\cite{argilla-feedback} hold over 1.5 billion characters. Translating this amount of  data for example with a commercial Translation API, like the DeepL Pro API\footnote{\url{https://www.deepl.com/en/pro/}}, would cost over 30.000 Euros.

Due to advances in the quality of open-source translation models, such costly APIs are not the only option. That's why, for this project, the ALMA model from Microsoft\cite{Xu2023-alma}, which has shown outstanding quality in translation tasks, was used to translate the data. 

To get a high throughput while maintaining quality, the vLLM library for inference of the ALMA Model \cite{vllm} has been used on a Nvidia A100 80GB GPU. The texts have been broken into chunks separated by the newline character to maintain the original text structure as best as possible. After the translation, the chucks were concatenated again. With this approach, the translation cost of these two datasets could be reduced to approximately 30 Euros, which is 100 times cheaper. This low price has been achieved because the University of Applied Science Ansbach has the necessary infrastructure. But even with a rented GPU, the costs would sum up to about 300 Euros, which still offers huge savings.

Concerning the datasets used, special care was taken to ensure that only data whose license also permits commercial use is included. This approach made it necessary to filter the Open-Platypus dataset from the LAION Team \cite{Plüster_2023}, as it contains data only approved for research. 

\subsection{Supervised Finetuning}

The process of supervised fine-tuning and alignment training is closely related to the process described in the paper for Zephyr model \cite{tunstall_zephyr_2023} and the Alignment Handbook by HuggingFace \cite{alignment_handbook2023}.
The base model has been swapped for the German adaption of the Mistral model\footnote{\url{https://huggingface.co/LeoLM/leo-mistral-hessianai-7b}}. The hyperparameters used for training on an 8x A100 80GB Instance can be seen in Table \ref{tab:hyper_sft}. For the SFT training data conversations with less than 2049 tokens have been selected from the previously mentioned instruction datasets.
% write about context length here

\begin{table}[h!]
    \centering
    \setlength{\abovecaptionskip}{6pt} % Adjust the spacing here

    \begin{minipage}{.45\textwidth}
        \centering
        \begin{tabular}{lr}
            \toprule
            Parameter & Value \\
            \midrule
            Batch Size & 512 \\
            Number of Steps & 228\\
            Packing & True\\
            Learning Rate Scheduler  & cosine \\
            Optimizer   &   Adam    \\
            Train Loss & 0.8767\\
            Eval Loss & 0.8753\\
            \bottomrule
        \end{tabular}
        \caption{Hyperparameters for SFT-Training}
        \label{tab:hyper_sft}
    \end{minipage}
    \hfill % Add horizontal space between the tables
    \begin{minipage}{.45\textwidth}
        \centering
        \begin{tabular}{lr}
            \toprule
            Parameter & Value \\
            \midrule
            Batch Size & 64 \\
            Number of Steps & 941 \\
            Learning Rate Scheduler & Linear \\
            Optimizer & Adam \\
            Train Accuracy & 78.75\% \\
            Eval Accuracy & 82.5\% \\
            \bottomrule
        \end{tabular}
        \caption{Hyperparameters for DPO-Training}
        \label{tab:hyper_dpo}
    \end{minipage}
    \label{table for paramters}
\end{table}

\subsection{Direct Preference Optimization}

As with the SFT step the DPO training and parameters in Table \ref{tab:hyper_sft} are aligned with the Alignment Handbook \cite{alignment_handbook2023}. The main difference can be seen in the number of training epochs. While the Zephyr model was trained for three epochs Phoenix was only trained for one. As the paper from the Zephyr model \cite{tunstall_zephyr_2023} shows, the improvement from additional DPO epochs in mt-bench score is only marginal. The DPO training data consists of the translated version of the Argilla Ultrafeedback dataset \cite{argilla-feedback}.
\begin{table}[h!]
    \centering
    \setlength{\abovecaptionskip}{6pt}
    \begin{tabular}{lr}
        \toprule
        Metric & Value \\
        \midrule
        First Turn & 6.39375 \\
        Second Turn & 5.1625 \\
        \midrule
        \multicolumn{2}{c}{Categories} \\
        \midrule
        Writing & 7.45 \\
        Roleplay & 7.9 \\
        Reasoning & 4.3 \\
        Math & 3.25 \\
        Coding & 2.5 \\
        Extraction & 5.9 \\
        STEM & 7.125 \\
        Humanities & 7.8 \\
        \midrule
        Average & 5.778125 \\
        \bottomrule
    \end{tabular}
    \caption{MT-Bench results of Phoenix}
    \label{tab:metrics}
\end{table}
\section{Evaluation}

As with training, the evaluation of German models is still an underdeveloped research area. Nevertheless, the LAION team \cite{Plüster_2023} has adapted the mt-bench and parts of the lm-evaluation harness \cite{eval-harness}. The results shown in Table \ref{tab:metrics} present the answer to the German mt-bench\footnote{\url{https://github.com/bjoernpl/FastEval}} of the Phoenix model. While the model still has some shortfalls in math and coding our model is able to outperform the Llama-2-70b-chat version of the LAION team\cite{Plüster_2023} in Reasoning and Roleplay and performs quite well compared to other models of the same size.

While this test alone is not sufficient for evaluating the full diversity of LLMs, it is one of the only available standardized sources for the German language. This displays the necessity of further research in multilingual LLM evaluation. To extend the comparison between the models, we compared the Mistral model of the LAION team (LeoLM-Mistral-7B-Chat)\cite{Plüster_2023} against Phoenix in the German versions of the HellaSwag, ARC and MMLU Benchmarks\footnote{\url{https://github.com/bjoernpl/lm-evaluation-harness-de/tree/mmlu_de}}.

\begin{table}[h]
    \centering
    \setlength{\abovecaptionskip}{6pt}
    \begin{tabular}{|c|c|c|c|}
        \hline
        \textbf{Model} & \textbf{HellaSwag-de} & \textbf{ARC-challenge-de}  &   \textbf{MMLU-DE} \\
        \hline
        LeoLM-Mistral-7B-Chat & 0.47639912 & 0.38310580 &  0.38944 \\
        \hline
        Phoenix & \textbf{0.5281} &  \textbf{0.4428} &    \textbf{0.3952} \\
        \hline
    \end{tabular}
    \caption{LM Evaluation Harness-DE results}
    \label{tab:performance}
\end{table}

%However, the comparison is not accurate because the Ultrafeedback Dataset from \cite{argilla-feedback} contains prompts from the TruthfulQA dataset. Since this information could not have been taken into account before training these results lack comparability.

\section{Conclusion}

In this paper we present, one of the first German DPO-aligned LLM. As the evaluation shows it performs on par with some of the best available models of its size and even manages to compete with 10 times larger models. Furthermore, we showcase how to create high-quality translated data at low cost to improve the LLM training process for different languages and task.

\clearpage

% \section*{Acknowledgments}
% This was was supported in part by......

%Bibliography
\printbibliography[heading=bibintoc, title={References}]

\newpage

\section{Appendix}
This section shows the comparison of the outputs for the same prompts between the Mixtral-8x7B Mixture of Experts(MoE) model and the Phoenix model with 7B parameters. For the comparison of the prompts side by side, we use the playground from COAI.

\begin{figure}[ht]
\centering
\includegraphics[width=\textwidth,keepaspectratio]{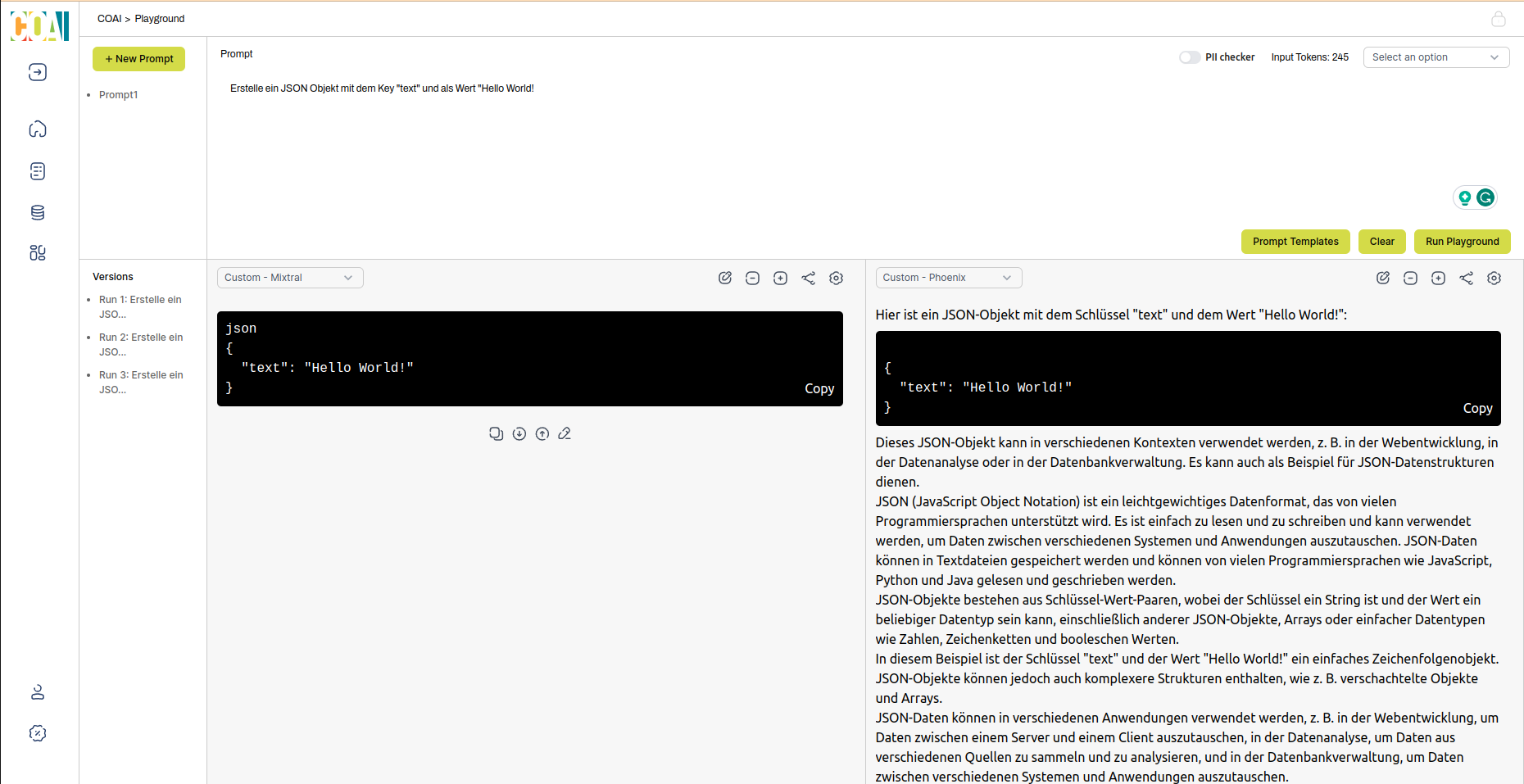}
\caption{The COAI Playground with the outputs for the two models side by side. The Mixtral MoE model is on the left and the Phoenix model is on the right.}
\label{fig:myimage}
\end{figure}

\begin{table}[ht]
\centering
\caption{Comparison of Responses from Two Different LLMs: Mixtral MoE and the Phoenix}
\label{tab:llm_comparison}
\begin{tabularx}{\textwidth}{>{\raggedright}p{3cm} >{\raggedright}X >{\raggedright\arraybackslash}X}
\toprule
\textbf{Prompt} & \textbf{Mixtral MoE} & \textbf{Phoenix} \\
\midrule
Erstelle ein JSON Objekt mit dem Key "text" und als Wert "Hello World!" & Generates only the JSON object & Generates the JSON object, and talks about the use of JSON objects, the concept of Key and Value. \\

\midrule
Wie hoch muss man einen Fisch an einer Angel hängen, damit eine Katze ihn nicht erreichen kann? & Talks about how the fishing line should be kept at a height. Talks about the distance of 1.5 to 2 meters over the floor. & Talks about how it is not possible to keep the fish safe from a cat, as the cats are skilled climbers. Talks about the life of the fish and its safety. \\
% Add more rows as needed
\bottomrule
\end{tabularx}
\end{table}

\end{document}